# Relative rationality: Is machine rationality subjective?

Tshilidzi Marwala

University of Johannesburg

South Africa

tmarwala@uj.ac.za

**Abstract:**

Rational decision making in its linguistic description means making logical decisions. In essence, a rational agent optimally processes all relevant information to achieve its goal. Rationality has two elements and these are the use of relevant information and the efficient processing of such information. In reality, relevant information is incomplete, imperfect and the processing engine, which is a brain for humans, is suboptimal. Humans are risk averse rather than utility maximizers. In the real world, problems are predominantly non-convex and this makes the idea of rational decision-making fundamentally unachievable and Herbert Simon called this bounded rationality. There is a trade-off between the amount of information used for decision-making and the complexity of the decision model used. This paper explores whether machine rationality is subjective and concludes that indeed it is subjective.

**Introduction**

Unbounded rational decision making is the concept of making decisions with perfect information, using a perfect brain in an optimized fashion (Bicchieri, 1993). Simply put, rational decision making is the optimal processing of complete information to achieve a goal. Because this whole process is optimized then it results in the maximization of utility. Just this statement makes unbounded rationality an impossible goal. Nobel Prize winner Herbert Simon identified that information is not perfect as it is often incomplete, imprecise and imperfect (Simon, 1982). He also observed that the human brain is not a perfect machine. Nobel Prize winner Kahneman realized that the human brain is not efficient in the sense that it does not maximize utility but it minimizes the risk (Kahneman, 2011). In essence humans are risk averse on making decisions. The optimization aspect is also not possible because of the impossibility of identifying the global optimum solution (Horst and Tuy, 1996). Marwala viewed this problem in two ways, first as a convex problem where a global optimum solution is accessed and it is known that it is an optimum solution (Nesterov, 2004). Second, in non-convex problems we cannot be sure that the identified solution is a global solution. This in a way is akin to number theory in mathematics where we do not know the highest prime number for sure (Hall, 2018). Herbert Simon termed these limitations bounded rationality.

As we are living in the era of intelligent machines such as artificial intelligence (AI), there are some big questions that we ought to ask (Marwala, 2004; Gidudu et. al., 2007; Vilakazi and Marwala, 2007). Are machines more rational than humans? (2) Can rationality be measured? (3) Is group rational decision-making more rational than individual decision making? (4) Is machine rationality subjective (Senter, 1977)? Marwala (2018a) studied whether AI machines are more rational than humans and concluded that even though machines and humans are both bounded rationally because of the trade-off between model complexity and model dimension as well as the limitation of optimization routines, intelligent machines are more rational than human beings. Furthermore, Marwala (2018b) studied whether rationality can be measured and concluded than rationality indeed can be measured. This paper studies whether rationality is subjective. To answer this question three crucial questions are posed. The first is whether a single goal optimization decision making process is subjective. The second is whether a multi-criteria optimization decision process is subjective. The third is whether the trade-off between model complexity and model dimension makes the information processing machine subjective. We then establish the linkages between the answers to these three questions to rationality. Based on these answers we conclude whether the conclusions observed make rationality subjective.

**Is a single goal optimization subjective?**

Rationality involves optimization. One cannot maximize utility without optimizing. If Thendo wanted to travel from Johannesburg to Cairo and there is a direct British Airways flight between the two cities but decides to fly from Johannesburg to London then to Cairo at twice the cost for no other reason except that he wanted to reach Cairo, then we will deem such a person irrational. This is because he is choosing a longer trip which costs twice as much without gaining any additional utility such as additional comfort (same airline). Travelling from Johannesburg to Cairo directly has higher utility than travelling from Johannesburg to London to Cairo at twice the cost on the same airline. Agents that choose lower utility over higher utility are not rational. How about if there is a Lufthansa flight from Johannesburg to Cairo at the same price as traveling between the two cities directly on a British Airways flight? If an agent deems that both these trips have the same utility, which one should he choose? Choosing a Lufthansa flight gives the exact same utility as choosing the British Airways flight. The concept of equal utility is illustrated in Figure 1. When two decisions have the same utility, in optimization language is called multiple optimum solutions (Xing et. al., 2010). In this regard, Thendo can choose Lufthansa flight because it sounds better than British Airways. This decision choice is subjective because it is based on other factors other than utility because utility is subjective. Therefore, the presence of multiple optimal solutions makes the decision choice subjective, and therefore in this case rationality is subjective. In essence, we define subjective machine decision as a decision which is not based on utility because the choices presented have equal utilities and therefore we cannot use utility to choose one decision over another. With decisions with identical utility, how does an intelligent machine makes a subjective choice? In such a scenario and in order to avoid a deadlock, the intelligent machine should make a random choice to make a subjective decision. These random choices which are not based on utility, because of identical utility values of available choices, are our definition of machine subjectivity.

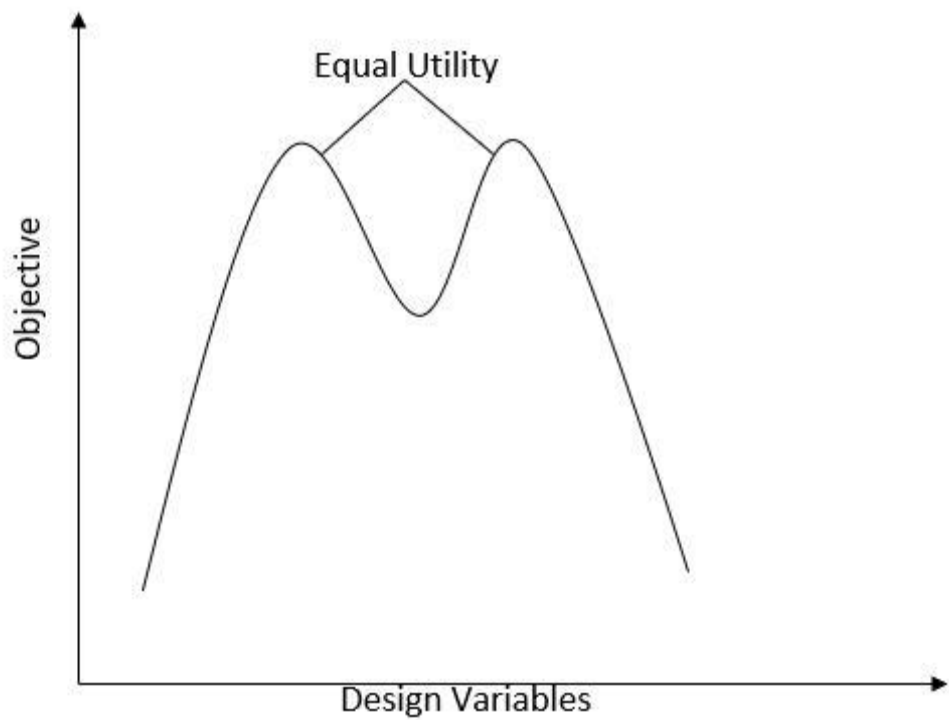

Figure 1 An illustration of an optimization process with two equal utilities

**Is multi-criteria optimization subjective?**

In aircraft design there is a trade-off between the strength of the aircraft and the fuel consumption. If we design the aircraft to be super strong using materials such as steel then it becomes too heavy and requires too much fuel to fly. If we use super-light material it becomes too fragile and might break apart mid-air and kill the passengers. So in this design problem, the goal is to maintain some level of structural strength but not too much to require lots of fuel to fly. Because the strength-fuel efficiency problem is in conflict, this is a multi-criteria optimization problem (Miettinen, 1999). The cost function of this problem is a combination of the strength of the aircraft, which should be maximized, and the fuel consumption, which should be minimized. How much weights should be assigned to the strength maximization and the fuel minimization is subjective. Pareto introduced a concept whereby a designer decides weights assignments of these two goals to form the overall goal and then optimize the overall goal. The different weights assignments and the corresponding overall optimal solutions form what is called the Pareto Optimality Frontier and this is shown in Figure 2 (Mathur, 1991). Where one ends up on this efficient frontier is completely subjective. If a rational decision making process involves optimizing a multi-criteria optimization problem, the solution has multiple solutions located at the Pareto frontier and therefore it is subjective.

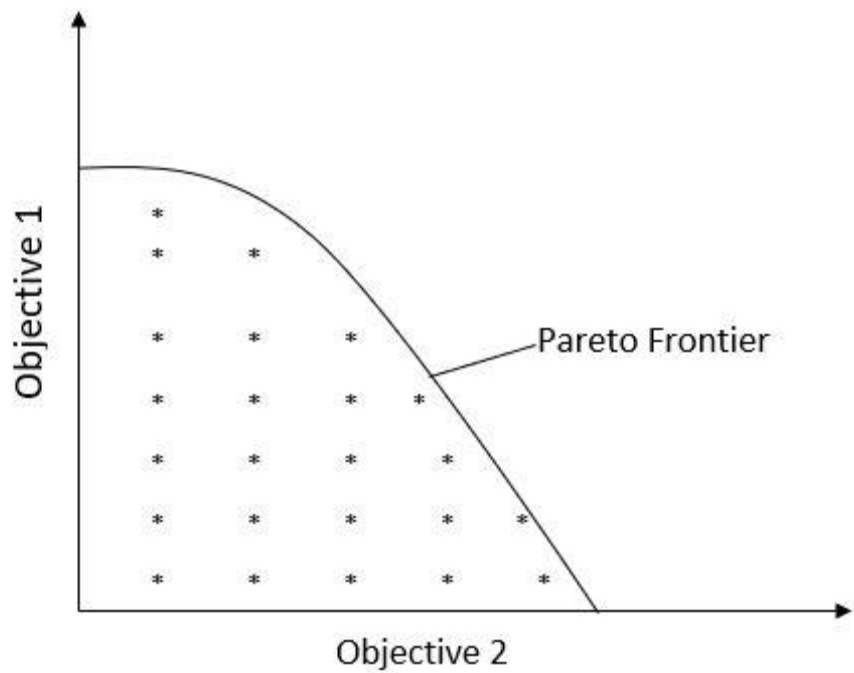

Figure 2 An illustration of the Pareto efficient frontier

**The curse of dimensionality and model complexity**

When we create artificial intelligence machines to make decisions the process involves using data that is used to train these machines. The data has a dimension. For example, if we are building a machine that uses the temperatures at 12:00 in the last two days to predict the temperature in the following day at 12:00, then this data set uses 2 variables to predict 1 variable. In this case, the dimension of the input data is 2 and that of the output is 1. When one builds this intelligent model, the complexity of the model is defined as the size of the free variables in the model. The practice in this modelling exercise is that the least complex model that describes the observed data is the most preferred one. This is called Occam's razor and is named after William of Occam (Gernert, 2007).

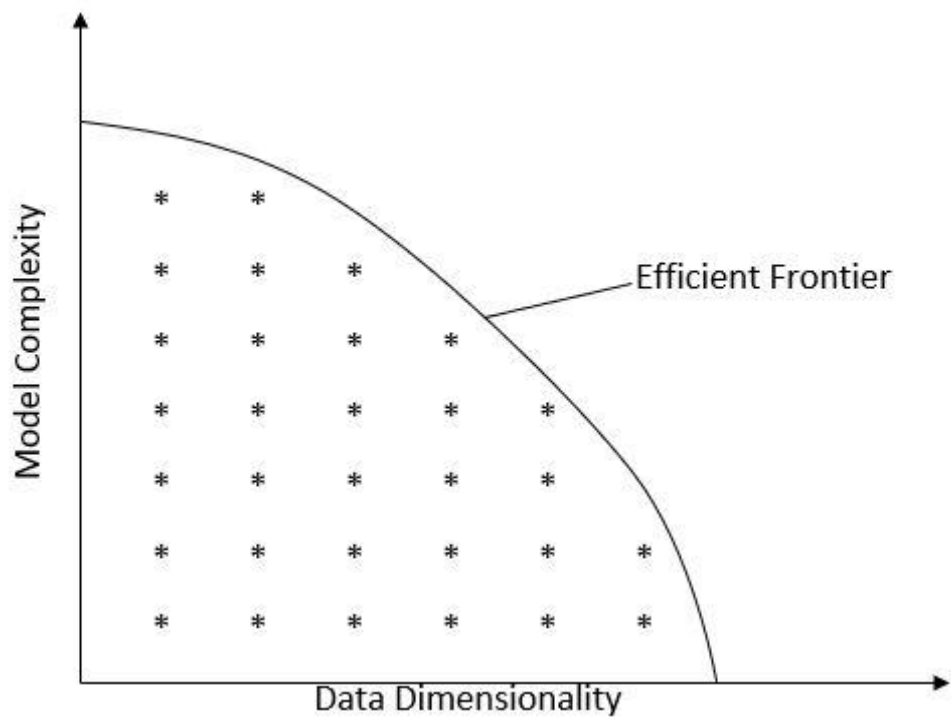

Figure 3 An illustration of the tradeoff between model complexity and data dimensionality

When we model a system, not all input variables are equally relevant. Some of them will be marginally relevant. Including more variables than needed often leads to a phenomenon called the curse of dimensionality (Bellman, 1957; Trunk, 1979). The inclusion of marginally relevant variables makes the model unnecessarily complex which compromises the effectiveness of the model. The trade-off between the dimension of the model and the complexity of the model gives an efficiency frontier which is illustrated in Figure 3. When constructing this intelligent model one needs to decide where in the efficiency frontier one needs to create a model, i.e. the chosen dimension for a desired complexity. This choice between model dimension and complexity is subjective. If this model is used for rational decision making, this model is bounded because the model is not perfect and furthermore the rationality of this model is subjective.

**Is machine rationality subjective?**

Machine rationality is subjective because of the three reasons and the first is that optimization, whether for single or multiple goals, is subjective. The second reason is that the trade-off between model dimension and complexity makes rationality subjective. This trade-off between dimension and complexity does not just occur when intelligent machines make decisions but also when human beings make decisions. When humans make decisions, the more variables they use to make decisions the less effective they are able to process such variables. The less variables they use to make decisions the better they are able to process the information. The choice that humans make on how many variables they use to process information versus the brain process effort which is a subjective choice ultimately makes bounded rational decisions subjective. Thirdly, training machines using data originating from subjective humans produces subjective machines.

**Conclusion**

This paper studied whether machine rationality is subjective. It identified the fact that optimization can sometimes lead multiple identical optimal utilities as one of the reasons why machine rationality is subjective. Secondly, it identifies the fact that some of the bounded rational decisions involve the use of multi-criteria optimization where the importance of each goal is subjective as a contributory factor to the subjectivity of machine rationality. Thirdly, it identifies the subjective trade-off between the dimension of the information used for decision making and the complexity of the model as the contributory factor to the subjectivity of machine rationality. It also identifies that machines that are trained with data that is subjective are subjective. In conclusion, this paper concludes that machine rationality is subjective.

**References:**

Bellman, R.E. (1957) Dynamic programming. Princeton University Press.

Bicchieri, C. (1993). Rationality and Coordination, New York: Cambridge University Press

Gernert, D. 2007. Ockham's Razor and its improper use. Journal of Scientific Exploration. 21, 135–140

Gidudu, A., Hulley, G. and Marwala, T. (2007) Classification of images using support vector machines. arXiv preprint arXiv:0709.3967

Hall, M. (2018) The Theory of Groups. Dover Books on Mathematics. Courier Dover Publications.

Horst, R. and Tuy, H. (1996) Global Optimization: Deterministic Approaches, Springer.

Kahneman, D. (2011) Thinking, Fast and Slow. Farrar, Straus and Giroux

Marwala, T. (2018a) The limit of artificial intelligence: Can machines be rational? arXiv:1812.06510.

Marwala, T. (2018b) Can rationality be measured? arXiv:1812.10144.

Marwala, T. (2004) Fault classification using pseudomodal energies and probabilistic neural networks. Journal of engineering mechanics 130 (11), 1346-1355

Mathur, V.K. (1991). "How well do we know Pareto optimality?". The Journal of Economic Education. 22 (2): 172–178.

Miettinen, K. (1999). Nonlinear Multi-objective Optimization. Springer.

Nesterov, Y. (2004). Introductory Lectures on Convex Optimization, Kluwer Academic Publishers

Senter, N. (1977) Is Rationality Radically Subjective? The Southwestern Journal of Philosophy, Vol. 8, No. 2, pp. 159-166Simon, H. (1982). Models of Bounded Rationality, Vols. 1 and 2. MIT Press

Trunk, G.V. (1979). "A Problem of Dimensionality: A Simple Example". IEEE Transactions on Pattern Analysis and Machine Intelligence. PAMI-1 (3): 306–307

Vilakazi, C.B. and Marwala, T. (2007) Online incremental learning for high voltage bushing condition monitoring. International Joint Conference on Neural Networks, 2521-2526

Xing, B., Gao, W.J., Nelwamondo, F.V., Battle, K. and Marwala, T. (2010) Ant colony optimization for automated storage and retrieval system. 2010 IEEE Congress on Evolutionary Computation (CEC), 1-7.